\documentclass{article}
\usepackage{spconf,amsmath,graphicx}

\usepackage{enumitem}
\setlist{nosep, leftmargin=14pt}

\usepackage{mwe} % to get dummy images
\usepackage{amssymb}
\usepackage{spconf,amsmath,graphicx,booktabs,multirow}
\usepackage{array}
\usepackage{xcolor}

\usepackage{graphicx} % for \scalebox

\title{Foundation Model-guided Iteratively Prompting and Pseudo-Labeling for Partially Labeled Medical Image Segmentation}
\name{Qiaochu Zhao$^{1,2}$, Wei Wei$^{2,3}$, David Horowitz$^{2,4}$ , Richard Bakst$^{5}$ , and Yading Yuan$^{2,3\dagger}$\thanks{$^{\dagger}$Corresponding author: yading.yuan@columbia.edu.}}
\address{
    $^{1}$Department of Biomedical Engineering, Columbia University, USA \\
    $^{2}$Department of Radiation Oncology, Columbia University Irving Medical Center, USA \\
    $^{3}$Data Science Institute, Columbia University, USA\\
    $^{4}$Herbert Irving Comprehensive Cancer Center, Columbia University, USA\\
    $^{5}$Department of Radiation Oncology, Icahn School of Medicine at Mount Sinai, USA
}
\begin{document}
%\ninept
%
\maketitle
\begin{abstract}
Automated medical image segmentation has achieved remarkable progress with \textbf{\emph{fully}} labeled data. However, site-specific clinical priorities and the high cost of manual annotation often yield scans with only a subset of organs labeled, leading to the \textbf{\emph{partially}} labeled problem that degrades performance. To address this issue, we propose IPnP, an Iteratively Prompting and Pseudo-labeling framework, for partially labeled medical image segmentation. IPnP iteratively generates and refines pseudo-labels for unlabeled organs through collaboration between a trainable segmentation network (specialist) and a frozen foundation model (generalist), progressively recovering full-organ supervision. On the public dataset AMOS with the simulated partial-label setting, IPnP consistently improves segmentation performance over prior methods and approaches the performance of the fully labeled reference. We further evaluate on a private, partially labeled dataset of 210 head-and-neck cancer patients and demonstrate our effectiveness in real-world clinical settings.
\end{abstract}
\begin{keywords}
Partially Labeled Segmentation, Prompt-based Learning, Pseudo-labels, Iterative Refinement
\end{keywords}
\section{Introduction}
\label{sec:intro}

\begin{figure}[t]
\centering
\begin{minipage}[b]{0.23\linewidth}
  \centering
  \includegraphics[width=\linewidth]{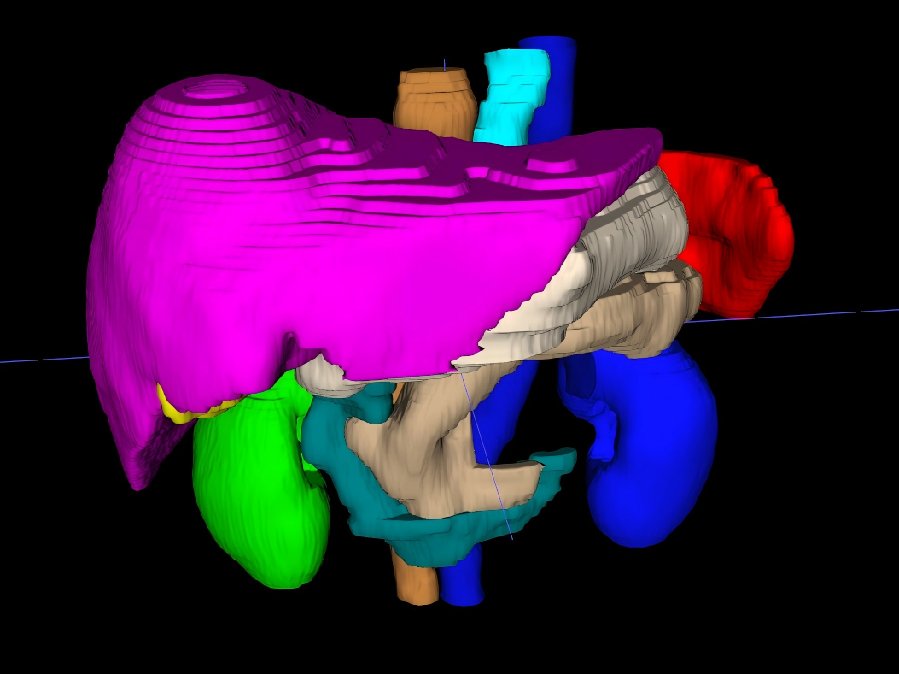}
  \centerline{\scriptsize (a) Full-labeled GT}
\end{minipage}
\hfill
\begin{minipage}[b]{0.23\linewidth}
  \centering
  \includegraphics[width=\linewidth]{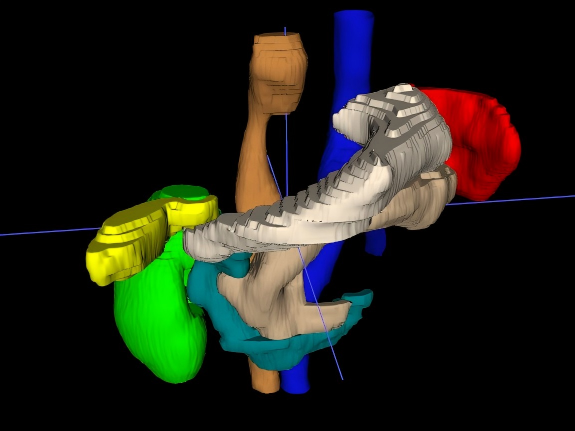}
  \centerline{\scriptsize (b) Partial-labeled GT}
\end{minipage}
\hfill
\begin{minipage}[b]{0.23\linewidth}
  \centering
  \includegraphics[width=\linewidth]{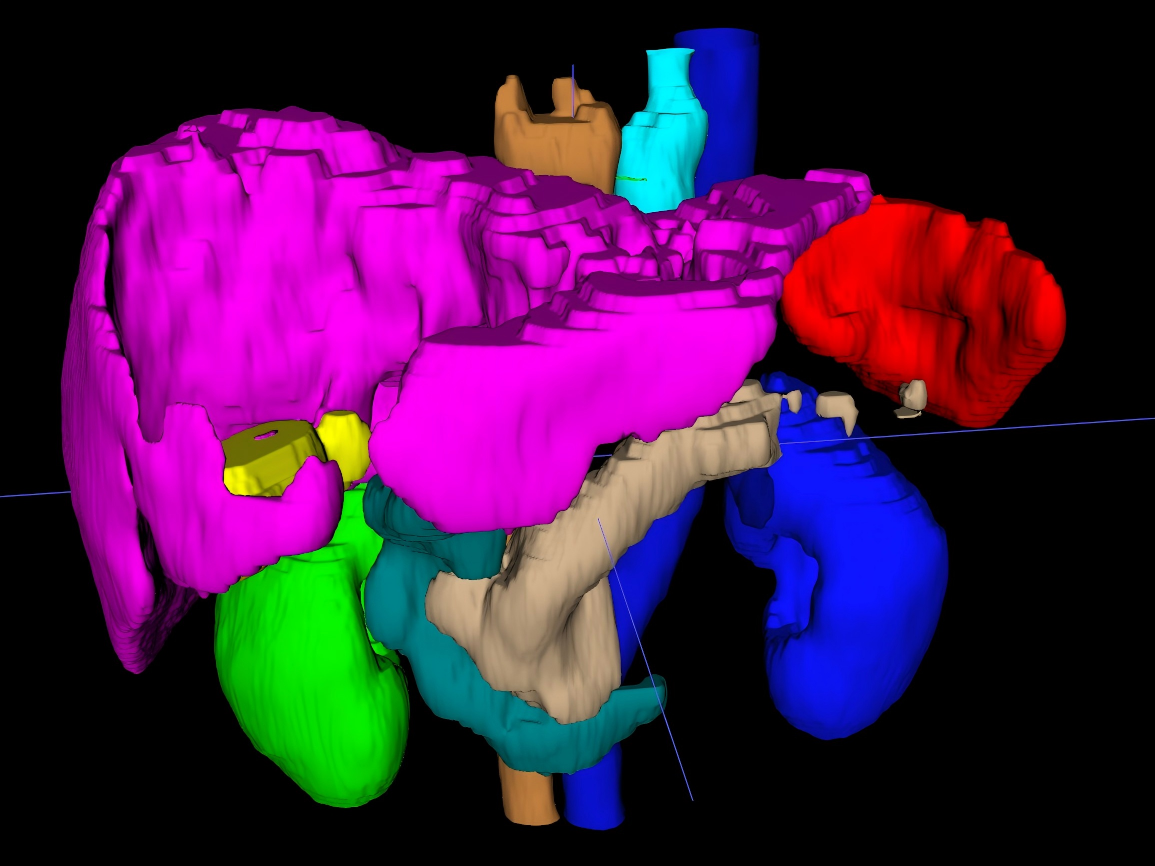}
  \centerline{\scriptsize (c) nnUNet result}
\end{minipage}
\hfill
\begin{minipage}[b]{0.23\linewidth}
  \centering
  \includegraphics[width=\linewidth]{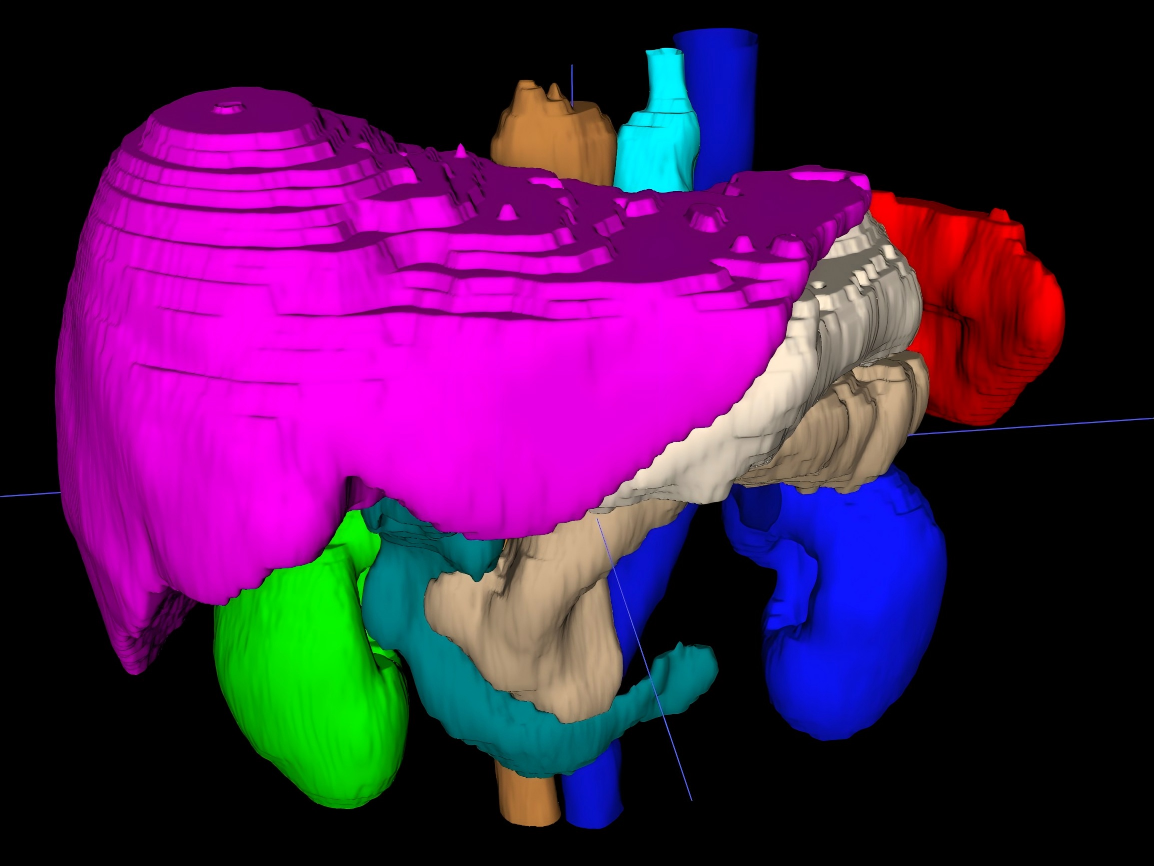}
  \centerline{\scriptsize (d) Our IPnP result}
\end{minipage}
\caption{
Visualization of partially supervised segmentation: (a) full reference mask (evaluation only); (b) partial-label mask as supervision in training; (c) Output of the nnU-Net baseline~\cite{Isensee2021}; (d) Output of the proposed IPnP framework.}
\label{fig:1}
\end{figure}

Accurate medical image segmentation is essential for clinical applications such as disease diagnosis and treatment planning. Although deep learning has achieved outstanding performance in multi-organ segmentation with fully annotated data~\cite{Ronneberger2015, Zhou2018, Cao2022, Isensee2021}, real-world clinical scenarios often diverge from this ideal setting. Varied clinical focus and the high cost of manual annotation result in only a subset of organs being labeled, as illustrated in Fig.~\ref{fig:1} (b). This motivates the need for segmentation methods under partially labeled conditions. 

To enable full-organ segmentation from partially labeled data, most existing methods supervise only the labeled organs and simply ignore the unlabeled ones~\cite{gonzalez2018multi,Fang2020,xie2023learning}. This reduces the effective supervisory signal and limits the learning of complete anatomical structures. Generating pseudo-labels from model predictions for unlabeled organs is a common remedy, but noise and error propagation often degrade performance~\cite{jin2025enhancing,li2023segment}. Recently, prompt-based medical segmentation foundation models~\cite{Wang2024, ma2025medsam2} have shown strong generalization and zero-shot capability, motivating a generalist–specialist strategy in which a foundation model (\emph{generalist}) generates pseudo-labels and a task-specific network (\emph{specialist}) learns from both ground-truth and pseudo supervision.

However, obtaining clean pseudo-labels from foundation models is challenging, as they typically rely on high-quality manual prompts.
To address this, we propose IPnP, a generalist–specialist framework for partially labeled segmentation.
A trainable specialist interacts with a frozen generalist to automatically generate and refine pseudo-labels for unlabeled organs, enabling full supervision from incomplete annotations.
Specifically, IPnP employs an iterative prompt and refinement scheme: the specialist is first trained on partial labels to generate coarse segmentation. High-confidence regions are then enclosed by bounding boxes, which serve as prompts for the generalist to produce candidate masks for unlabeled organs. The resulting masks are refined via confidence-based filtering and fed back to subsequent training, progressively improving supervision quality and organ structural integrity. To further suppress noise in pseudo-labels, we introduce a voxel selection loss that restricts supervision to reliable voxels. Consequently, IPnP improves segmentation accuracy and structural integrity over baseline, as shown in Fig.~\ref{fig:1}(c-d). To sum up, our contributions are as follows:

\begin{itemize}
    \item We propose an iterative framework that progressively generates and refines pseudo-labels for unlabeled organs for partially label medical image segmentation.
    \item We introduce a voxel-level selection loss that suppresses unreliable voxels and noise in pseudo-labels.
    \item Our method substantially improves performance on the simulated partial-labeled AMOS dataset and approaches the referenced fully labeled upper bound. Evaluation on a 210-case head-and-neck clinical dataset further demonstrates its strong generalization.
\end{itemize}

\section{METHODS}
\label{sec:format}

\subsection{Overview}

\begin{figure}[t]
  \centering
  \includegraphics[width=0.8\linewidth]{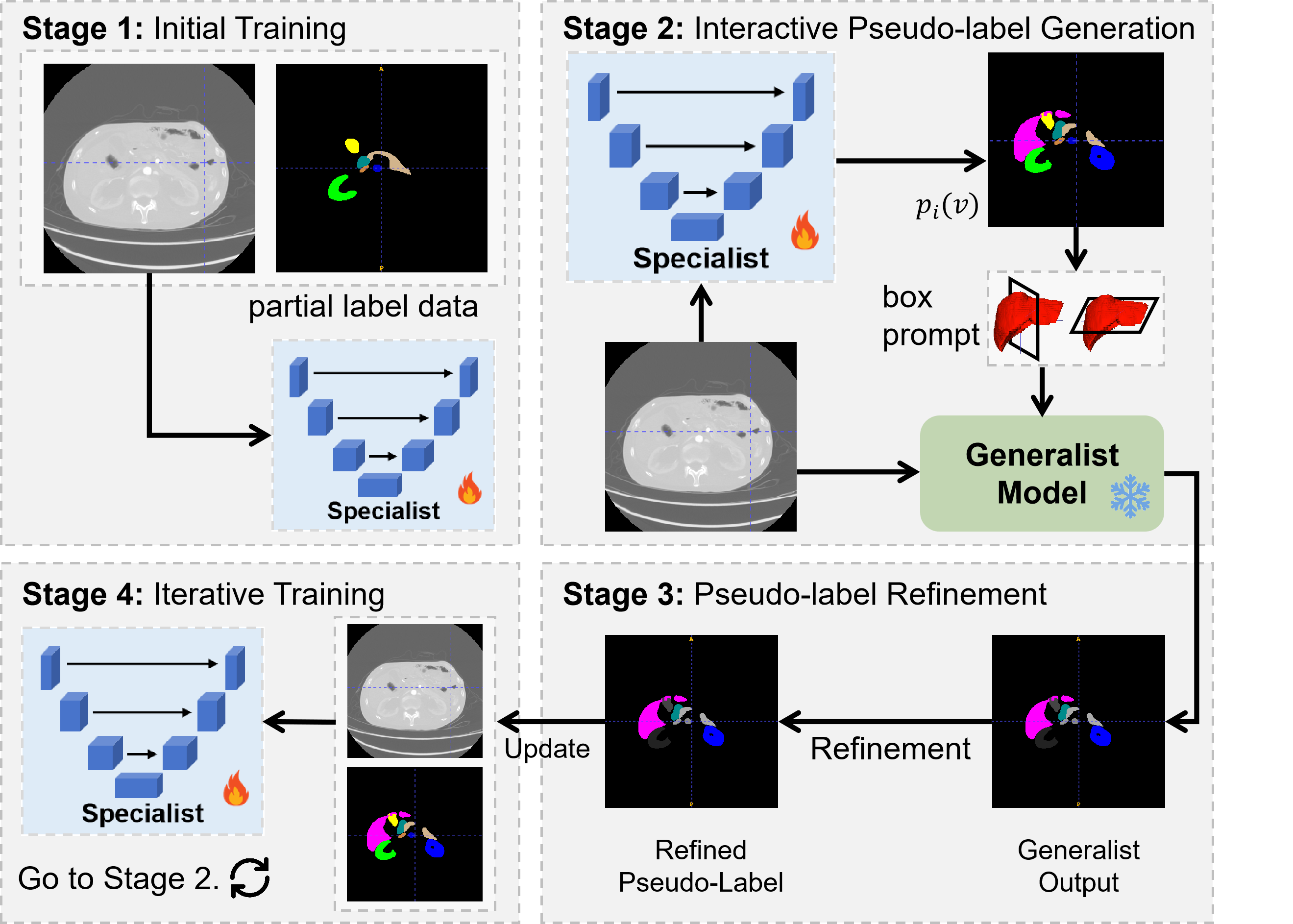}
  \caption{Overview of the proposed IPnP framework.}
  \label{fig:2}
\end{figure}

IPnP comprises four stages:
(1) Initial Training: the specialist is trained on partial label data to obtain initial segmentation ability;
(2) Interactive Pseudo-label Generation: specialist outputs are converted into box prompts that guide the generalist to generate pseudo-labels;
(3) Pseudo-label Refinement: unreliable voxels are filtered by confidence and entropy;
(4) Iterative Training: refined pseudo-labels are incorporated to retrain the specialist with voxel-level selection loss.

\subsection{Initial Training}

The specialist is first trained on partial label organs to obtain a basic
segmentation capability. The resulting model produces coarse but informative
predictions, which are converted into box prompts to guide the generalist in
subsequent iterations.

\subsection{Interactive Pseudo-label Generation}

Given a volume $X \in \mathbb{R}^{H \times W \times D}$, the generalist 
$f_{\theta}$ outputs voxel logits $\mathbf{P} \in \mathbb{R}^{C \times H \times W \times D}$,
and the predicted label map $\hat{Y}$ is obtained by a voxel-wise argmax.
Let $\mathcal{L}$ and $\mathcal{U}$ denote the sets of labeled and unlabeled organs, 
respectively, where $\mathcal{L} \cup \mathcal{U}=\{1,\dots,C\}$. 
For each unlabeled class $i \in \mathcal{U}$, we define a 3D mask $M_i$ as the
voxel set $\{v \mid \hat{Y}(v)=i\}$.

From \(M_i\), we extract the axial and sagittal median slices (\(z_{\text{med}}\) and \(x_{\text{med}}\)) that contain foreground voxels. On each slice, a 2D bounding box for class \(i\) is computed from the foreground coordinates and expanded by a fixed padding \(p\):
\begin{equation}
B_{*,i} = \operatorname{pad}\!\left(\operatorname{bbox}\!\left(M_i^{(*_{\text{med}})}\right),\, p\right),
\quad * \in \{z, x\}.
\end{equation}

The two orthogonal boxes $(B_{z,i}, B_{x,i})$ are jointly used as \textit{box prompts} to the generalist model, where the orthogonal views provide spatial information of the organ within the volume. The model takes image volume $X$ and both box prompts as input and predicts the pseudo-label $\tilde{M}_i$ for organ $i$.

This interaction allows the specialist to guide the generalist by providing box prompts that localize the target organ. Leveraging its zero-shot ability, the generalist then produces a pseudo-label for the unlabeled organ.

\subsection{Pseudo-label Refinement}
\label{sec:2.4}

Three constraints are used to filter unreliable voxels in the pseudo-labels before producing the final pseudo-labels.

\textbf{(1) Class probability threshold.}
For each voxel $v$, the softmax probability of class $i$ is denoted as $p_i(v)$. Only voxels satisfying $p_i(v) \geq \tau_\text{cls}$ are kept in the pseudo-label, where $\tau_\text{cls}$ is a fixed confidence threshold.

\textbf{(2) ROI constraint.}
A 3D ROI is constructed by combining the two box prompts $(B_{z,i}, B_{z,i})$ and dilating them by a margin $\delta_\text{roi}$. Voxels outside the ROI are removed:
\begin{equation}
\tilde{M}_i(v) = 0, \quad \text{if } v \notin \operatorname{ROI}(B_{z,i}, B_{x,i}, \delta\text{roi}).
\end{equation}

\begin{figure}
    \centering
    \includegraphics[width=0.95\linewidth]{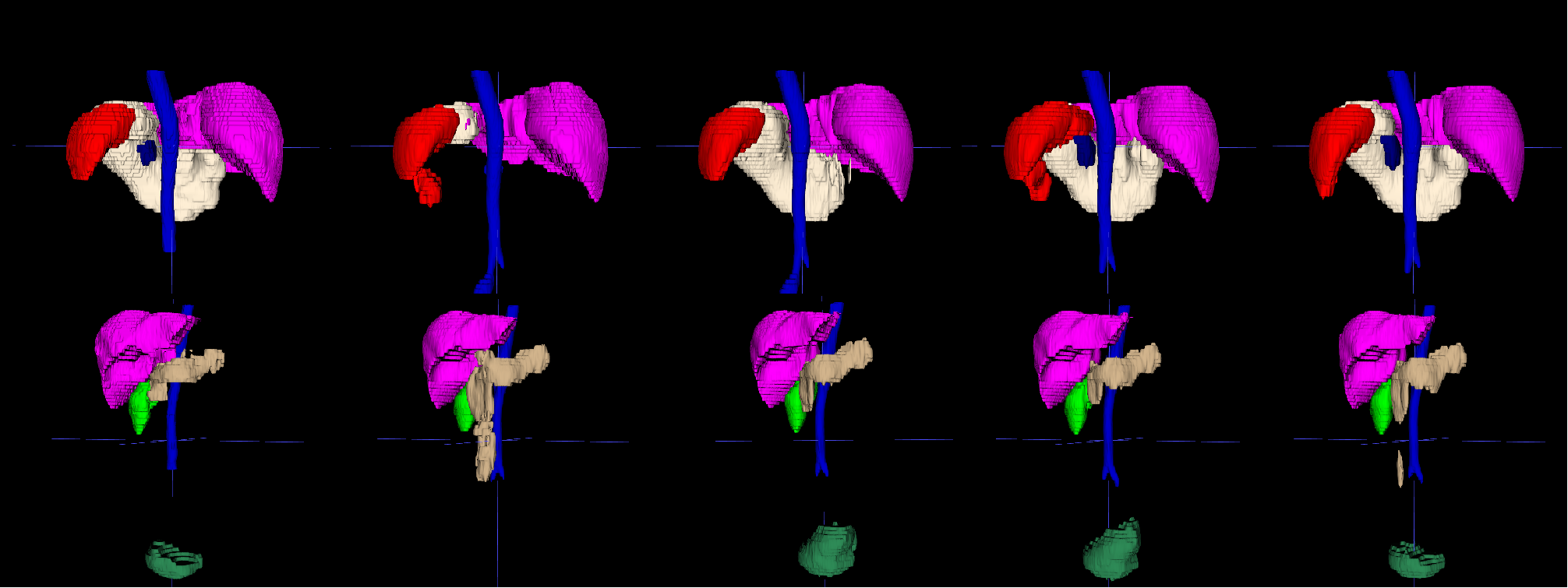}
    \caption{Qualitative visualization of pseudo-labels in the IPnP framework. The first column shows the ground truth, while columns 2–5 show the pseudo-labels generated after epoch 50, epoch 150, epoch 350, and epoch 450, respectively.}
    \label{fig:3}
\end{figure}

\textbf{(3) Entropy constraint.} 
For each organ $i$, the voxel-wise entropy $H(v)$ is computed from class probabilities $p_c(v)$:
\begin{equation}
\bar{H}_i = \frac{1}{|\tilde{M}_i|} \sum_{v \in \tilde{M}_i} H(v),
\quad H(v) = - \sum_{c=1}^{C} p_c(v) \log p_c(v).
\end{equation}
Let $\bar{H}_i^{(t)}$ and $\bar{H}_i^{(t-1)}$ denote the mean entropy of the current and previous iterations. The pseudo-label $\tilde{M}_i^{(t)}$ is updated only when the mean entropy decreases: $\bar{H}_i^{(t)} < \bar{H}_i^{(t-1)}$.
Entropy serves as an uncertainty measure: low entropy indicates confident predictions, whereas high entropy reflects noisy or unstable voxels.
To illustrate the progressive refinement achieved by our iterative training, Fig.~\ref{fig:3} shows pseudo-labels produced at different training epochs.

\subsection{Iterative Training}

IPnP is trained in an iterative scheme that alternates between pseudo-label generation and specialist re-training. The specialist’s predictions are converted into box prompts that guide the generalist to produce pseudo-labels. These pseudo-labels are then refined using the constraints in Section~\ref{sec:2.4}. The entropy constraint is activated only in later iterations, allowing early pseudo-labels to expand organ regions despite higher uncertainty, and becoming more effective once the model predictions stabilize. The refined pseudo-labels are added to training set in each iteration, then the specialist is re-trained.

To further suppress noise during optimization, we employ a voxel-level selection (VLS) strategy inspired by learning with noisy labels~\cite{Xu2022}. For each voxel $v$, let $p_i(v)$ denote the predicted probability of class $i$, $\hat{Y}(v)=\arg\max_i p_i(v)$ the predicted label, and $y(v)$ the supervision target. For an image $s$, let $\mathcal{P}_s$ be the set of pseudo-labeled classes. A voxel-wise selection mask is defined as
\begin{equation}
m(v)=
\begin{cases}
\mathbb{I}[\hat{Y}(v)=y(v)], & y(v)\in\mathcal{P}_s,\\[3pt]
1, & y(v)\notin\mathcal{P}_s,
\end{cases}
\end{equation}
so that only reliable pseudo-labeled voxels contribute to the loss, while all
truly labeled voxels remain fully supervised. The VLS mask is applied to both CE and Dice losses.

\begin{table*}[t]
\centering
\caption{Segmentation performance of IPnP and other methods on the AMOS dataset, reported in DSC (\%) and HD95 (mm).}
\label{tab:table1}

\scriptsize
\setlength{\tabcolsep}{3.0pt}
\renewcommand{\arraystretch}{0.95}
\resizebox{0.98\textwidth}{!}{
\begin{tabular}{|l|cc|cccc|cccc|cccc|cccc|}
\hline
\multirow{3}{*}{\textbf{Organ}} &
\multicolumn{2}{c|}{\shortstack{\textbf{Full Supervision}\\(Full Labels)}} &
\multicolumn{4}{c|}{\shortstack{\textbf{Full Supervision}\\(Partial Labels)}} &
\multicolumn{4}{c|}{\shortstack{\textbf{Partial Supervision}\\(Partial Labels)}} &
\multicolumn{4}{c|}{\shortstack{\textbf{TransDoDNet}\\(Partial Labels)}} &
\multicolumn{4}{c|}{\shortstack{\textbf{IPnP}\\\textbf{(Ours)}}} \\
\cline{2-19}
& \multicolumn{2}{c|}{\textbf{Full labels}} &
\multicolumn{2}{c}{\textbf{67\% label}} & \multicolumn{2}{c|}{\textbf{33\% label}} &
\multicolumn{2}{c}{\textbf{67\% label}} & \multicolumn{2}{c|}{\textbf{33\% label}} &
\multicolumn{2}{c}{\textbf{67\% label}} & \multicolumn{2}{c|}{\textbf{33\% label}} &
\multicolumn{2}{c}{\textbf{67\% label}} & \multicolumn{2}{c|}{\textbf{33\% label}} \\
\cline{2-19}
& \textbf{DSC} & \textbf{HD95} &
\textbf{DSC} & \textbf{HD95} & \textbf{DSC} & \textbf{HD95} &
\textbf{DSC} & \textbf{HD95} & \textbf{DSC} & \textbf{HD95} &
\textbf{DSC} & \textbf{HD95} & \textbf{DSC} & \textbf{HD95} &
\textbf{DSC} & \textbf{HD95} & \textbf{DSC} & \textbf{HD95} \\
\hline
Spleen             & 97.04 & 2.79 & 94.42 & 11.43 & 59.39 & 56.32 & 95.57 & 56.41 & 83.62 & 398.96 & 96.32 & 99.97 & 93.16 & 219.32 & \textbf{96.64} & \textbf{3.01} & \textbf{95.87} & \textbf{3.74} \\
Right Kidney       & 96.49 & 34.91 & 95.88 & \textbf{8.34} & 67.62 & 67.69 & 95.42 & 97.57 & 91.66 & 146.15 & 96.26 & 19.62 & \textbf{91.64} & 15.66 & \textbf{96.45} & 15.74 & 89.73 & \textbf{10.89} \\
Left Kidney        & 96.66 & 13.57 & 93.95 & 3.39 & 65.78 & \textbf{33.44} & 96.23 & 40.43 & \textbf{92.43} & 92.20 & \textbf{96.55} & 18.07 & 88.45 & 221.13 & 96.36 & \textbf{3.02} & 92.17 & 157.51 \\
Gall Bladder       & 82.96 & 11.74 & 81.46 & \textbf{19.18} & 55.34 & 55.70 & 80.64 & 38.73 & 58.59 & 622.84 & \textbf{85.90} & 60.49 & 68.93 & 279.13 & 82.02 & 19.66 & \textbf{78.86} & \textbf{23.04} \\
Esophagus          & 85.90 & 6.62 & 83.24 & \textbf{9.14} & 66.46 & 47.32 & \textbf{85.13} & 15.00 & 44.02 & 688.45 & 84.20 & 8.61 & 47.75 & 239.98 & 83.93 & 10.68 & \textbf{78.20} & \textbf{15.08} \\
Liver              & 97.44 & 7.56 & 92.85 & 9.16 & 53.60 & 101.45 & \textbf{97.14} & 19.14 & 81.98 & 287.37 & 95.68 & 45.69 & 95.59 & 75.83 & 97.03 & \textbf{7.72} & \textbf{96.17} & \textbf{11.10} \\
Stomach            & 92.14 & 9.51 & 81.44 & 32.98 & 43.32 & 119.43 & 87.45 & 50.39 & 77.74 & 251.09 & 71.67 & 34.38 & 84.54 & 165.25 & \textbf{90.63} & \textbf{18.26} & \textbf{87.07} & \textbf{35.23} \\
Aorta              & 95.52 & 14.43 & 94.07 & 4.62 & 48.13 & 54.39 & 94.53 & 28.82 & 91.80 & 100.09 & 94.45 & 18.94 & 86.09 & 61.04 & \textbf{94.97} & \textbf{4.49} & \textbf{93.30} & \textbf{8.43} \\
IVC & 92.04 & 3.24 & 87.99 & 3.68 & 60.77 & 30.21 & \textbf{91.68} & 3.79 & 84.33 & 319.38 & 89.68 & 2.90 & 78.65 & 162.15 & 90.96 & \textbf{2.83} & \textbf{87.38} & \textbf{4.63} \\
Pancreas           & 88.10 & 5.44 & 81.62 & 21.56 & 45.57 & 97.42 & \textbf{87.47} & 13.66 & \textbf{83.40} & 53.07 & 86.79 & 30.31 & 48.18 & 103.37 & 87.24 & \textbf{11.99} & 82.04 & \textbf{8.16} \\
Right Adrenal      & 78.90 & 4.24 & 76.03 & 13.70 & \textbf{69.15} & 36.27 & \textbf{77.38} & 10.46 & 70.49 & 220.08 & 76.04 & 11.41 & 33.15 & 42.45 & 76.17 & \textbf{8.75} & 66.65 & \textbf{6.55} \\
Left Adrenal       & 80.20 & 5.95 & 77.01 & 6.99 & \textbf{72.18} & 32.47 & \textbf{78.12} & 44.24 & 57.42 & 612.53 & 77.73 & 11.48 & 21.74 & 158.03 & 77.83 & \textbf{6.85} & 70.10 & \textbf{9.22} \\
Duodenum           & 83.07 & 13.18 & 76.35 & 26.98 & 56.90 & 56.87 & \textbf{81.50} & \textbf{13.38} & 72.67 & 154.96 & 78.99 & 24.76 & 65.13 & 102.37 & 79.97 & 20.19 & \textbf{73.51} & \textbf{33.75} \\
Bladder            & 88.33 & 7.48 & 84.20 & 13.98 & 53.52 & 35.12 & 67.86 & 97.62 & 69.87 & 278.45 & \textbf{88.69} & 136.94 & 42.66 & 133.90 & 88.31 & \textbf{12.66} & \textbf{80.55} & \textbf{25.60} \\
Prostate           & 83.42 & 8.37 & 81.19 & \textbf{9.11} & 72.34 & 30.84 & 79.76 & 31.56 & 10.78 & 143.15 & 67.47 & 34.94 & 62.27 & 82.64 & \textbf{82.36} & 13.99 & \textbf{75.42} & \textbf{26.55} \\
\hline
\textbf{Average}   & 89.21 & 9.95 & 85.45 & 12.92 & 59.34 & 57.96 & 86.39 & 36.70 & 71.39 & 295.89 & 85.76 & 35.99 & 67.19 & 137.74 & \textbf{88.06} & \textbf{10.53} & \textbf{83.12} & \textbf{25.32} \\
\hline
\end{tabular}}
\end{table*}

\section{EXPERIMENTS}
\label{sec:pagestyle}

\begin{figure}
    \centering
    \includegraphics[width=0.95\linewidth]{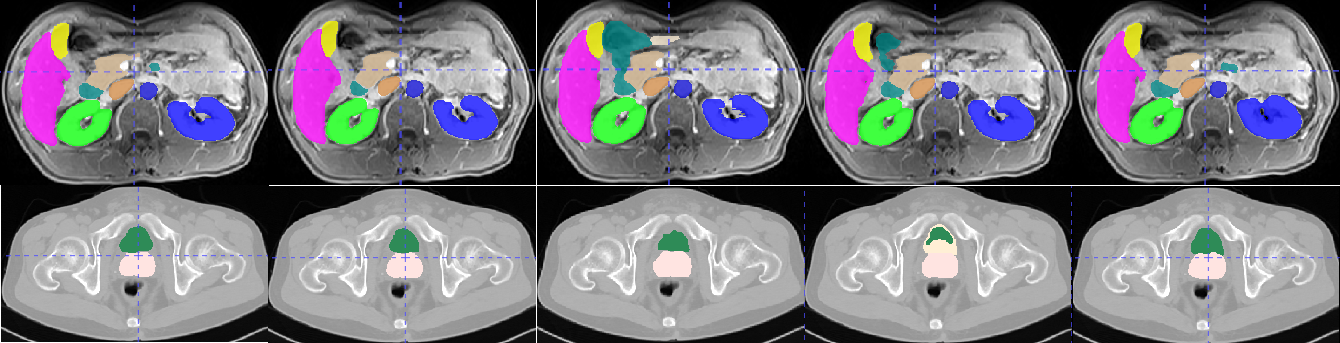}
    \caption{Qualitative comparison between methods. The first column shows the ground truth, and the second to fifth columns correspond to Full Supervision (Full Labels), Full Supervision (Partial Labels), Partial Supervision (Partial Labels), and IPnP (Partial Labels), respectively.}
    \label{fig:4}
\end{figure}

\begin{figure*}
  \centering
  \includegraphics[width=0.74\textwidth]{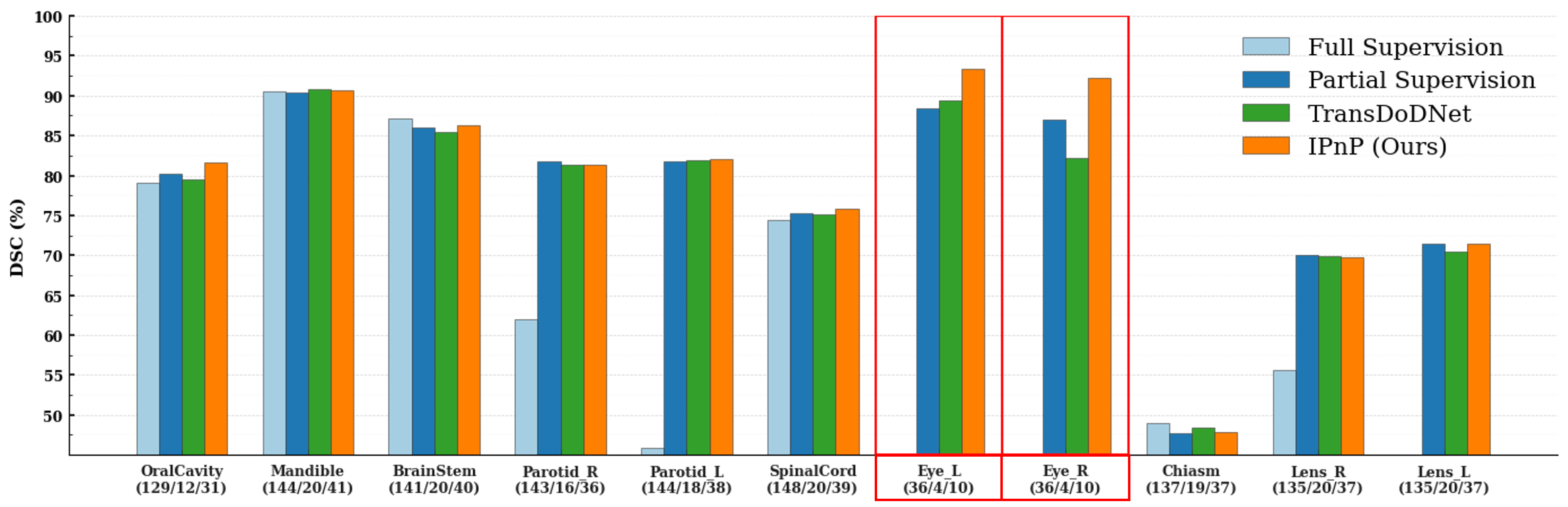}
  \caption{Segmentation performance of IPnP and other methods on HnN dataset, reported in DSC (\%). The numbers in parentheses indicate the counts of labeled organs in the training, validation, and testing sets, respectively, for each organ.}
  \label{fig:HnN}
\end{figure*}

\subsection{Dataset}

First, we performed simulated partial-label segmentation experiments on the AMOS abdominal multi-organ dataset \cite{Ji2022}, which provides full annotations for 15 organs. To simulate partial-label scenarios, we randomly removed annotations from each scan, keeping 67\% and 33\% organ labels for moderate and challenging settings, respectively. The dataset was split into 192 training, 48 validation, and 120 testing cases.

To further evaluate the generalization ability of IPnP in real-world clinical scenarios, we applied it to our private dataset with 210 head-and-neck (HnN) cancer patients who received radiotherapy from the Department of Radiation Oncology at Mount Sinai Hospital (New York, NY, USA). The organs at risk (OARs) include oral cavity, mandible, brain stem, parotids (left/right), spinal cord, eyes (left/right), optic nerves (left/right), chiasm and lenses (left/right).  The dataset was split into 148 training, 21 validation, and 41 testing cases.

\subsection{Implementation Details}

The specialist was based on nnUNetv2~\cite{Isensee2021} and trained for 1000 epochs with SGD, a linearly decayed learning rate (from 0.01), and a batch size of 2. The generalist for pseudo-labeling was nnInteractive~\cite{Isensee2025}, a promptable segmentation model with zero-shot ability. Box padding, class-probability threshold, and ROI dilation were set to \(p=6\), \(\tau_{\text{cls}}=0.4\), and \(\delta_{\text{roi}}=3\). All experiments were performed on NVIDIA RTX 6000 Ada GPUs and evaluated by Dice Similarity Coefficient (DSC) and 95th-percentile Hausdorff Distance (HD95).

\subsection{Results}

We compare the proposed IPnP framework with one fully-labeled upper bound reference, two relative baselines, and one recently state-of-the-art partial-label segmentation method: (1) Full supervision (full labels): trained on the fully annotated dataset using a standard multi-class loss, serving as the upper bound reference;
(2) Full supervision (partial labels): trained on the partially labeled dataset with the same multi-class loss, representing the lower bound due to incorrect supervision from missing organs;
(3) Partial supervision (partial labels): trained on the partially labeled dataset using a partially labeled loss, where each organ channel is activated by a sigmoid instead of softmax. Binary cross-entropy and Dice losses are computed for each labeled organ;
(4) TransDoDNet~\cite{xie2023learning} (partial labels):  Trained on the partially labeled dataset using the task-specific dynamic head that generates convolutional parameters based on annotated organs. TransDoDNet is a strong baseline that has been commonly used as a representative method for comparison in partial-label segmentation studies.
Methods (2) and (3) represent two intuitive ways of handling unlabeled organs: they either treat them as background or simply ignore them, which respectively introduces wrong supervision or discards potentially useful supervision.

We report the quantitative results on the partial-labeled AMOS dataset in Table~\ref{tab:table1}, and show qualitative comparisons in Fig.~\ref{fig:4}.
Overall, IPnP consistently outperforms baselines (2), (3) and (4) under both 67\% and 33\% labeled settings. When trained with full supervision loss, the average DSC drops from 89.2 (fully labeled) to 85.8 (67\% labeled) and 59.3 (33\% labeled), showing the adverse impact of incorrectly treating missing organs as background. The partially labeled loss and TransDoDNet's dynamic head mitigate this issue; however, their DSC and HD95 remain suboptimal, indicating inaccurate segmentation and unstable boundary. In contrast, IPnP achieves 88.1 / 83.1 DSC and 10.53 / 25.32 mm HD95, demonstrating both higher overlap and better boundary precision,  approaching the full labels upper bound and demonstrating strong robustness to severe annotation incompleteness.

We analyze the effect of the proposed VLS loss. Without VLS, the average DSC drops from 88.06 to 86.46, with notable degradations in challenging organs such as the esophagus (83.93 $\rightarrow$ 81.92), and adrenal glands (right: 76.17 $\rightarrow$ 72.73; left: 77.83 $\rightarrow$ 71.67), indicating that VLS effectively suppresses noisy and improves segmentation accuracy.

We further evaluate IPnP on the private head-and-neck dataset. As shown in Fig.~\ref{fig:HnN}, IPnP performs similarly to partial supervision and TransDoDNet in well-annotated organs, as expected under strong supervision. 
In contrast, for organs with limited annotations, IPnP benefits from generalist to recover missing labels. The left and right eyes each have only 36 labeled cases, where full supervision fails to segment (10.49 / 8.16 DSC). IPnP improves DSC from 88.5 and 89.4 to 93.4 for left eye, and from 86.9 and 82.2 to 92.2 for right eye, outperforming partial supervision and TransDoDNet.

\section{CONCLUSION}
\label{sec:typestyle}

In this work, we present IPnP, a generalist-guided iterative prompting and pseudo-labeling framework for partial label medical segmentation. With collaborative learning between a trainable specialist and a frozen generalist, IPnP generates, refines, and integrates pseudo-labels for unlabeled organs. With confidence-based refinement and a voxel-level selection loss that filters unreliable supervision, IPnP effectively suppresses noise and approaches full labels performance without additional manual annotations. Experimental results on both AMOS and clinical head-and-neck datasets demonstrate the robustness and generalization capability of IPnP under various degrees of incomplete annotation. Future work will include broader baseline comparisons, evaluation on additional datasets, and exploration of diverse prompting strategies to further improve robustness under partial-label settings.

\vfill
\pagebreak

\section{Compliance with ethical standards}
\label{sec:ethics}

This study was performed in line with the principles of the Declaration of Helsinki. The use of internal data was granted by the Institutional Review Board (IRB) of the Icahn School of Medicine at Mount Sinai (IRB-18-01242) and the IRB of Columbia University (IRB-AAAU5788). This research was also conducted retrospectively using human subject data made available in open access by the AMOS dataset (distributed under the CC BY-NC-SA license and hosted on the AWS Open Data platform). Ethical approval was not required as confirmed by the license attached with the open access data.

\section{Acknowledgments}
\label{sec:acknowledgments}

This work is supported by R21EB030209 from the National Institute of Biomedical Imaging and Bioengineering of the National Institutes of Health, USA. The content is solely the responsibility of the authors and does not necessarily represent the official views of the National Institutes of Health. This research has been partially funded through the generous support of Herbert and Florence Irving/the Irving Trust.

\bibliographystyle{IEEEbib}
\bibliography{strings,refs}

\end{document}